\documentclass[runningheads]{llncs}

\usepackage[T1]{fontenc}
\usepackage{graphicx,verbatim}
\usepackage[utf8]{inputenc}
\usepackage{amsmath}
\usepackage{amssymb}
\usepackage{bm}
\usepackage{multirow}
\usepackage{bbm}
\usepackage{booktabs}
\usepackage[misc]{ifsym}
\usepackage{hyperref}
\usepackage{color}

\begin{document}

\title{Hierarchical Text-Guided Brain Tumor Segmentation via Sub-Region-Aware Prompts}
\titlerunning{Hierarchical Text-guided Segmentation via Sub-Region-Aware Prompts}

\author{
Bahram Mohammadi\inst{1}\and
Ta Duc Huy\inst{2} \and
Afrouz Sheikholeslami\inst{1} \and \\
Qi Chen\inst{2} \and
Vu Minh Hieu Phan\inst{2} \and
Sam White\inst{2} \and
Minh-Son To\inst{3} \and \\
Xuyun Zhang\inst{1} \and
Amin Beheshti\inst{1} \and
Luping Zhou\inst{4} \and
Yuankai Qi\inst{1} 
}
\authorrunning{B. Mohammadi et al.}
\institute{
Macquarie University, Sydney, NSW, Australia \\
\email{mohammadibahram71@gmail.com} \and
Adelaide University, Adelaide, SA, Australia \and
Flinders University, Adelaide, SA, Australia \and
University of Sydney, Sydney, NSW, Australia
}

\maketitle

\begin{abstract}
Brain tumor segmentation remains challenging because the three standard sub-regions, i.e., whole tumor (WT), tumor core (TC), and enhancing tumor (ET), often exhibit ambiguous visual boundaries. Integrating radiological description texts with imaging has shown promise. However, most multimodal approaches typically compress a report into a single global text embedding shared across all sub-regions, overlooking their distinct clinical characteristics. We propose TextCSP (text-modulated soft cascade architecture), a hierarchical text-guided framework that builds on the TextBraTS baseline with three novel components: (1) a text-modulated soft cascade decoder that predicts WT$\to $TC $\to$ET in a coarse-to-fine manner consistent with their anatomical containment hierarchy. (2) sub-region-aware prompt tuning, which uses learnable soft prompts with a LoRA-adapted BioBERT encoder to generate specialized text representations tailored for each sub-region; (3) text-semantic channel modulators that convert the aforementioned representations into channel-wise refinement signals, enabling the decoder to emphasize features aligned with clinically described patterns. Experiments on the TextBraTS dataset demonstrate consistent improvements across all sub-regions against state-of-the-art methods by $1.7\%$ and $6\%$ on the main metrics Dice and HD95.


\end{abstract}

\section{Introduction}
Deep learning has made substantial progress on brain tumor segmentation, with architectures such as U-Net~\cite{olaf_2015_unet}, nnU-Net~\cite{isensee_2021_nnu}, and Swin-UNETR~\cite{hatamizadeh_2021_swin} achieving strong performance on the BraTS benchmarks~\cite{shi2025textbrats}. However, these methods operate exclusively on imaging data, ignoring the rich clinical context available in radiological reports. In practice, radiologists produce free-text annotations that describe the tumor's location, size, enhancement pattern, surrounding edema, and relationship to eloquent structures, which informs the distinction between sub-regions. For instance, the phrase "\textit{ring-enhancing mass with central necrosis}" immediately constrains the spatial relationship between ET and the non-enhancing core, a distinction that is often ambiguous from imaging alone.

The recently introduced TextBraTS~\cite{shi2025textbrats} benchmark provides the first publicly available dataset that pairs volumetric brain MRI with radiological text descriptions, together with a baseline model that fuses SwinTransformer features and BioBERT~\cite{lee_2020_biobert}
embeddings via cross-attention. 
This benchmark demonstrates the value of incorporating clinical language into tumor segmentation. Building on this idea, several text-guided segmentation frameworks have emerged 
~\cite{rokuss2025voxtell,liu2025medsam3,shi2025textbrats,luo2025vividmed,xin2025text3dsam,xie2026tvpnet}.
VoxTell~\cite{rokuss2025voxtell} maps free-form prompts to volumetric masks through multi-stage vision–language fusion, achieving strong zero-shot performance. In~\cite{xin2025text3dsam}, natural language descriptions are integrated into a SAM-inspired decoder for general 3D medical segmentation. TVPNet~\cite{xie2026tvpnet} leverages text–vision prompt interactions to better highlight small or clinically significant structures. Despite this progress, current approaches exhibit two key limitations for multi-region tumor segmentation. First, most models use a single shared output head, ignoring the anatomical containment hierarchy (ET $\subset$ TC $\subset$ WT) and often producing inconsistent predictions, such as ET appearing outside TC.
Second, they typically compress the entire report into a single global text embedding, which overlooks the fact that WT, TC, and ET depend on different clinical cues (e.g., edema vs. necrosis vs. enhancement). 

To address these limitations, we propose \textbf{TextCSP}, a unified hierarchical text-guided framework for brain tumor segmentation.
First, to exploit the anatomical containment hierarchy, we design a text-modulated soft cascade with three prediction branches: WT generates a soft spatial attention prior that guides TC, and TC in turn constrains ET, explicitly encoding the structure ET $\subset$ TC $\subset$ WT. Second, to overcome the limitation of the single global text embedding, TextCSP introduces sub-region-aware prompt tuning, where distinct soft prompts steer a LoRA-adapted BioBERT encoder to produce specialized text representations for WT, TC, and ET. Third, each branch further incorporates a text-semantic channel modulator, which injects its branch-specific text representation into decoder features through a squeeze-and-excitation mechanism, enabling channel-wise refinement aligned with clinically described patterns. These components together form a parameter-efficient yet expressive multi-modal architecture. 
Our main contributions are summarized as follows:
\begin{itemize}
    \item 
    We introduce a text-modulated soft cascade decoder with three independent prediction branches that encodes the anatomical containment hierarchy (ET $\subset$ TC $\subset$ WT), where coarser predictions guide finer sub-regions and text modulators inject essential information at each cascade level.

    \item 
    We propose sub-region-aware prompt tuning, in which distinct learnable soft prompts are prepended to a shared LoRA-adapted BioBERT encoder to produce three branch-specialized text representations.

    \item 
    The results on the TextBraTS dataset demonstrate consistent enhancement over the baseline. Specifically, average Dice and HD95 are improved by $1.7\%$ and $6\%$, respectively.
\end{itemize}

\begin{figure*}[t!]
\centering
\includegraphics[width=0.99\linewidth]{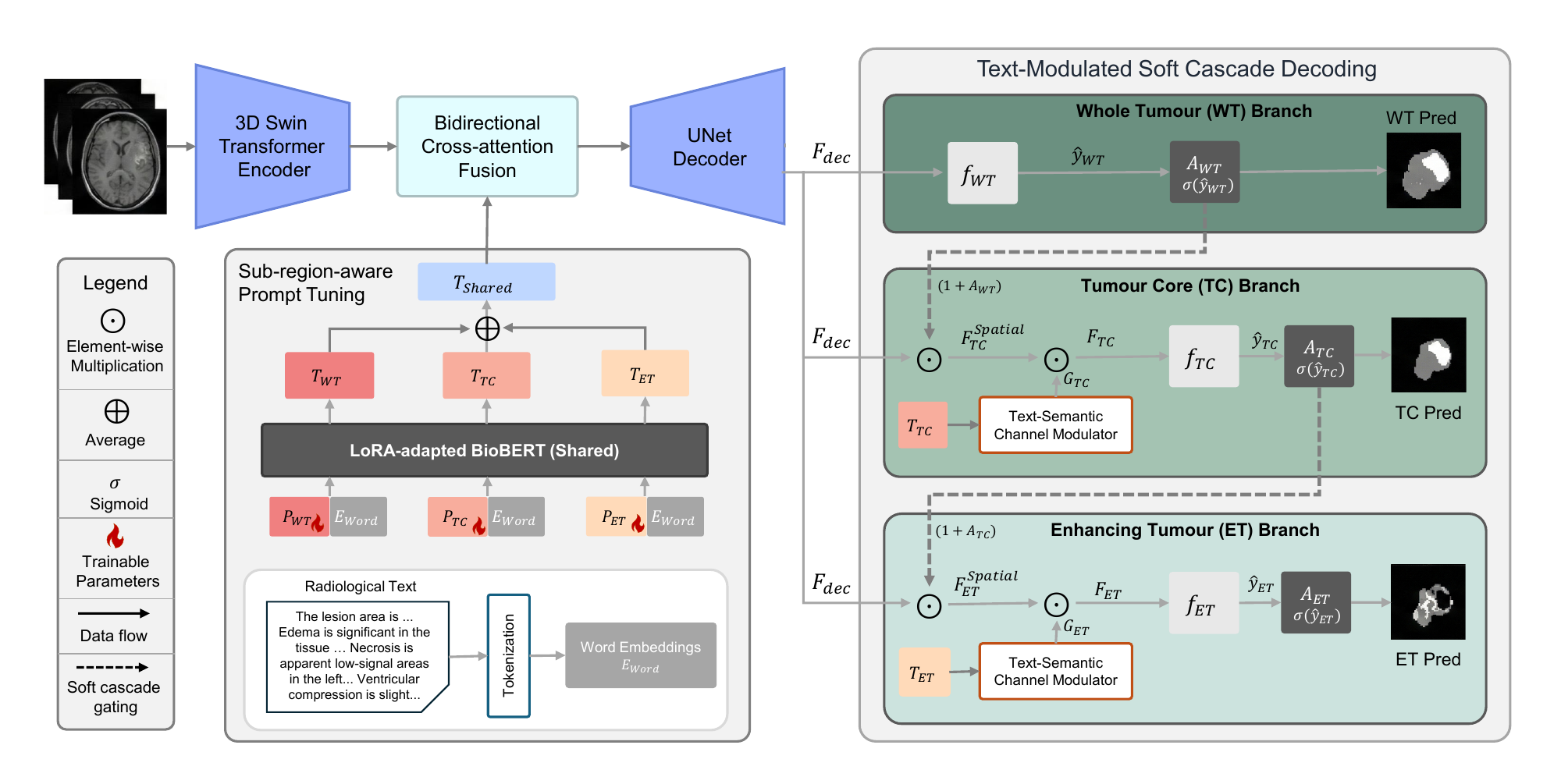}
   \vspace{-2mm}
   \caption{Main architecture of our TextCSP, which contains three key components: (1) a text-modulated soft cascade decoder that predicts WT$\to$TC$\to$ET in a coarse-to-fine manner (Sec.~\ref{sec::soft_cascade}); (2) sub-region-aware prompt tuning (Sec.~\ref{sec::prompt_tuninig}), which uses learnable soft prompts with a LoRA-adapted BioBERT encoder to generate specialized text representations tailored to each sub-region; and (3) text-semantic channel modulators that convert these representations into channel-wise refinement signals, enabling the decoder to emphasize features aligned with clinically described patterns~\ref{sec::text_modulation}.}
\label{fig::outline}
\end{figure*}

\section{Method}

We propose {TextCSP}, a unified hierarchical text-guided framework for brain tumor segmentation (Fig.~\ref{fig::outline}). Built upon the TextBraTS baseline, TextCSP introduces three novel components.
First, we design a \emph{soft cascade decoding strategy}  with three separate sub-region prediction heads (Sec.~\ref{sec::soft_cascade}) that explicitly enforce the anatomical containment hierarchy (ET $\subset$ TC $\subset$ WT). Second, we design a \emph{Sub-Region-Aware Text Adaptation}, which combines parameter-efficient fine-tuning (LoRA~\cite{hu_2021_lora}) with sub-region-aware prompt tuning to produce branch-specialized text representations, overcoming the limitation of a single global embedding (Sec.~\ref{sec::sub-region-text-adaptation}). Third, we design \emph{text-derived semantic channel modulators} that inject branch-specific linguistic priors into each cascade branch
(Sec.~\ref{sec::text_modulation}).

\vspace{-3mm}
\subsection{Baseline Architecture}
The baseline TextBraTS framework consists of a 3D Swin Transformer encoder that extracts hierarchical visual features at five resolutions from four-channel MRI volumes (T1, T1ce, T2, FLAIR), 
a text encoder from BioBERT that produces embeddings of radiological description text, 
a sequential cross-attention fusion module at the encoder bottleneck that enables bidirectional reasoning between image and text, 
and a U-Net decoder with skip connections that reconstructs high-resolution feature maps. 
These features are passed to a single output head predicting all three tumor sub-regions. We design three new components, detailed below, that largely improve this baseline.

\subsection{Soft Cascade Decoding with Separate Sub-Region Heads}
\label{sec::soft_cascade}
The baseline model may fail to exploit the anatomical hierarchy among brain tumor sub-regions because of using a single shared output head. We propose a soft cascade of three independent prediction heads, where coarser regions generate a spatial attention map for finer sub-regions. Let $F_{dec} \in \mathbb{R}^{B \times C \times D \times H \times W}$ denote the high-resolution decoder output features, where $C = 48$. The cascade proceeds through three sequential branches, each equipped with its own convolutional output head:

\vspace{1mm}
\noindent \textbf{WT Branch.}
The WT branch operates directly on the decoder features to predict the binary distinction between tumor and healthy tissue:
$
\hat{y}_{WT} = f_{WT}(F_{dec})
$,
$
A_{WT} = \sigma(\hat{y}_{WT})
$,
where $f_{WT}$ is the WT output head and $\sigma(.)$ is the sigmoid function. The resulting soft attention map $A_{WT} \in \mathbb{R}^{B \times 1 \times D \times H \times W}$ provides a voxel-wise probabilistic prior over the tumor location, which is propagated to the subsequent TC branch.

\vspace{1mm}
\noindent \textbf{TC and ET Branch.}
The TC branch incorporates spatial guidance from the WT prediction. The decoder features are modulated by the WT attention gate to focus computation on the predicted tumor region:
$
F_{TC}^{spatial} = F_{dec} \odot (1 + A_{WT}),
$
where $\odot$ denotes element-wise multiplication. The term $(1 + A_{WT})$ implements a soft residual gating mechanism: voxels within the predicted tumor region receive amplified features (up to $2\times$ when $A_{WT} \to 1$), while voxels outside the tumor retain their original features ($1\times$ when $A_{WT} \to 0$). This avoids information loss inherent in hard masking while still providing a strong spatial inductive bias.

The spatially gated features are then subject to text-based semantic modulation (described in Sec.~\ref{sec::text_modulation}), resulting in the final TC features, from which the TC logit is predicted:
$
F_{TC} = F_{TC}^{spatial} \odot G_{TC}$,
$
\hat{y}_{TC} = f_{TC}(F_{TC}).
$
The TC attention gate is then computed as $A_{TC} = \sigma(\hat{y}_{TC})$ and propagated to the next stage, \textit{i.e.}, ET branch. The ET logits $\hat{y}_{ET}$ are predicted in the same manner as in the previous stage.

\subsection{Sub-Region-Aware Text Adaptation}
\label{sec::sub-region-text-adaptation}
To overcome the limitation of a single shared text embedding, we propose \emph{Sub-Region-Aware Text Adaptation}. It comprises two components detailed below.

\vspace{-5mm}
\subsubsection{Parameter-Efficient Text Encoding via LoRA}
\label{sec::lora}
We propose using LoRA to fine-tune the text encoder, which learns task-specific representations that better capture the discriminative semantics required for sub-region differentiation. We apply LoRA specifically to the query and value projection matrices. 
The key projections and all feed-forward layers remain frozen, resulting in only a small number of trainable parameters. For a pretrained weight matrix $W \in \mathbb{R}^{d \times d}$, the adapted forward pass is $Wx + \frac{\alpha}{r}$, where $r$ is the rank and $\alpha$ is the scaling factor.

\vspace{-5mm}
\subsubsection{Sub-Region-Aware Prompt Tuning}
\label{sec::prompt_tuninig}
Each tumor sub-region relies on various linguistic cues: edema descriptors for WT, necrosis-related terms for TC, and enhancement patterns for ET. 
Hence, we introduce sub-region-aware prompt tuning, a lightweight mechanism that utilizes a single shared LoRA-adapted BioBERT model to produce three branch-specialized representations.
We define three sets of learnable soft prompt embeddings $\mathbf{P}_{\text{WT}}$, $\mathbf{P}_{\text{TC}}$, $\mathbf{P}_{\text{ET}}$ $\in$ $\mathbb{R}^{K\times d}$, with prompt length $K=4$ and hidden dimension $d=768$. For each sub-region $s~\in~\{\text{WT},\text{TC},\text{ET}\}$, the prompts are prepended to word embeddings and processed via the LoRA-adapted encoder:
$
        \mathbf{T}_s = \text{BioBERT}_{\text{LoRA}}\!\bigl(\mathbf{E}_s,\mathbf{M}_s\bigr)\bigl[\,:,\;K:\,\bigr],
$
where $\mathbf{M}_s$ is the extended attention mask,  $\mathbf{E}_s = [\,\mathbf{P}_s;\mathbf{E}_{\text{word}}\,]$,  $\mathbf{E}_{\text{word}}~\in~\mathbb{R}^{L\times d}$ contains the tokenized word embeddings, and the slicing discards prompt-position outputs.
This text encoder is shared across the three forward passes, and only the prompt embeddings differ, acting as lightweight steering vectors that condition the same encoder to produce semantically distinct outputs. 
The shared signal, $\mathbf{T}_{\text{shared}} = (\mathbf{T}_{\text{WT}} + \mathbf{T}_{\text{TC}} + \mathbf{T}_{\text{ET}})/3$, is used by the bottleneck cross-attention, while the branch-specific $\mathbf{T}_{\text{TC}}$ and $\mathbf{T}_{\text{ET}}$ are individually routed to their respective cascade branches for text-semantic channel modulation (Sec.~\ref{sec::text_modulation}).

\subsection{Text-Semantic Channel Modulation}
\label{sec::text_modulation}
While the WT boundary is largely determinable from visual appearance alone, the distinction between TC and ET sub-regions is highly semantic. Radiological texts frequently contain critical discriminative cues that directly inform which tissue subtypes are present and should be segmented. We exploit this observation by introducing text modulators that translate linguistic priors into channel-wise feature refinement signals. 
Each modulator follows the Squeeze-and-Excitation (SE) block paradigm~\cite{hu2019squeezeandexcitationnetworks}, but we replace the conventional spatial squeeze with a textual squeeze that derives global context from the radiological reports. 
Specifically, the contextualized token embeddings from the LoRA-adapted BioBERT are first aggregated via global average pooling across the token dimension to obtain a sentence-level embedding. This sentence embedding is then passed through a multi-layer perceptron (MLP) to produce channel-wise modulation weights $G = \sigma(W_2 \cdot \text{ReLU}(W_1 \cdot \bar{t}))$, where $W_1 \in \mathbb{R}^{(C/r) \times 768}$ projects from the text dimension to a compressed bottleneck of dimension $C/r = 12$, and $W_2 \in \mathbb{R}^{C \times (C/r)}$ projects back to the feature dimension $C = 48$. The sigmoid output $G \in \mathbb{R}^{B \times C}$ is reshaped to $(B, C, 1, 1, 1)$ and broadcast across the spatial dimensions for element-wise multiplication with the feature map.

Two independent modulator instances are instantiated, one for TC ($G_{TC}$) and one for ET ($G_{ET}$). No text modulator is applied to the WT branch, reflecting the design principle that the tumor-vs-background distinction is primarily a visual task, while sub-region differentiation benefits from semantic guidance.

\section{Experiments}

\subsection{Experimental Settings}

\noindent\textbf{Dataset.}
We evaluate TextCSP on the TextBraTS benchmark~\cite{shi2025textbrats}. Each data sample comprises four co-registered MRI sequences (T1, T1ce, T2, FLAIR) with voxel-wise annotations for WT, TC, and ET, paired with a free-text radiological description of tumor characteristics. We follow the official train and test splits provided by the benchmark.

\noindent\textbf{Evaluation Metrics.}
We adopt two main metrics per sub-region: \textbf{Dice}~\cite{dice_1945_measures} measures volumetric overlap, and 95th percentile Hausdorff Distance (\textbf{HD95})~\cite{huttenlocher_2002_comparing} measures boundary accuracy as the 95th percentile of symmetric surface distances (in mm), providing robustness to outlier errors compared to the standard Hausdorff distance. Both metrics are computed separately for WT, TC, and ET and averaged to obtain overall scores.

\noindent\textbf{Implementation Details.}
All experiments are conducted on a single NVIDIA RTX A6000 GPU using PyTorch~\cite{paszke_2017_pytorch} with MONAI~\cite{cardoso_2022_monai}. Input volumes are resized to $128\times128$ in width and height, and intensity-normalized per channel on nonzero voxels. For LoRA~\cite{hu_2021_lora}, we set rank $r=8$ and scaling factor $\alpha=16$ applied to the query and value projections of BioBERT~\cite{lee_2020_biobert}. The prompt length is $K=4$ tokens per sub-region. The text-semantic channel modulators use a reduction ratio of $r=4$. We train for $200$ epochs using Sharpness-Aware Minimization (SAM)~\cite{foret_2021_sam} with SGD~\cite{nesterov_1983_sgd} as the base optimizer (learning rate 0.1, momentum 0.9), combined with a linear warmup cosine annealing schedule (50 warmup epochs).

\begin{table*}[t!]
    \def\arraystretch{1}
    \setlength{\tabcolsep}{7pt}
    \caption{Comparison of TextCSP with the state-of-the-art methods on the TextBraTS dataset. $\dagger$ denotes the results reproduced on the same platform as ours.
    }
    \vspace{-2mm}
    \centering
        \resizebox{\textwidth}{!}{
        \begin{tabular}{lcccccccc}
            \toprule
            
            \multicolumn{1}{c}{\multirow{2}{*}{Methods}} & \multicolumn{4}{c}{Dice (\%) $\uparrow$} & \multicolumn{4}{c}{HD95 $\downarrow$} \\
            \cmidrule(r{2pt}){2-5} \cmidrule(l{2pt}){6-9} &
            \multicolumn{1}{c}{ET} & \multicolumn{1}{c}{WT} & \multicolumn{1}{c}{TC} & \multicolumn{1}{c}{Avg.} &
            \multicolumn{1}{c}{ET} & \multicolumn{1}{c}{WT} & \multicolumn{1}{c}{TC} & \multicolumn{1}{c}{Avg.} \\

            \hline

            3D-UNet~\cite{olaf_2015_unet} & 80.4 & 87.3 & 81.6 & 83.1 & 6.11 & 10.51 & 8.93 & 8.17 \\
            
            nnU-Net~\cite{isensee_2021_nnu} & 82.2 & 87.5 & 82.6 & 84.1 & \underline{4.27} & 11.90 & 8.52 & 8.23 \\
            
            SegResNet~\cite{hsu_2021_brain} & 80.9 & 88.4 & 82.3 & 83.8 & 6.18 & 7.28 & 7.41 & 6.95 \\

            Swin UNETR~\cite{hatamizadeh_2021_swin} & 81.0 & 89.5 & 80.8 & 83.8 & 5.95  & 8.23 & 7.03 & 7.07 \\
            
            Nestedformer~\cite{xing_2022_nestedformer} & 82.6 & 89.5 & 80.2 & 84.1 & 5.08 & 10.51 & 8.93 & 8.17 \\

            TextBraTS~\cite{shi2025textbrats} & \underline{83.3} & \underline{89.9} & \underline{82.8} & \underline{85.3} & 4.58 & \underline{5.48} & \textbf{5.34} & \underline{5.13} \\

            \hline
        
            TextBraTS$^{\dagger}$ & 82.8 & 89.6 & 82.5 & 84.9 & 5.28 & 8.59 & 6.77 & 6.88 \\

            \textbf{TextCSP} (Ours) & \textbf{85.3} & \textbf{90.7} & \textbf{85.1} & \textbf{87.0} & \textbf{3.95} & \textbf{4.98} & \underline{5.51} & \textbf{4.81} \\
        
            \bottomrule
            
        \end{tabular}}
    \vspace{-2mm}
    \label{tab::main_results}
\end{table*}

\subsection{Comparison with SOTA methods} 
Table~\ref{tab::main_results} compares TextCSP against the SOTA methods on the TextBraTS dataset. According to this table, TextCSP achieves the highest average Dice score of $87.0\%$, surpassing the best previously reported result, TextBraTS, by $1.7\%$. The improvement is consistent across all three sub-regions. Notably, the largest gain is observed for TC ($+2.6\%$), the sub-region most dependent on distinguishing necrotic and non-enhancing components from enhancing tissue. In terms of boundary accuracy, TextCSP achieves the best average HD95 of $4.81$ mm, improving upon TextBraTS by $\approx6\%$ ($0.32$ mm). The noticeable performance gap between TextBraTS and TextCSP highlights the advantage of moving from a shared textual representation to subregion-aware prompting. Rather than providing the same linguistic signal to all prediction targets, TextCSP adapts the text encoding to the discriminative requirements of each sub-region, leading to consistent enhancements across both overlap- and boundary-based metrics.

\begin{table}[t]
    \centering
    \setlength{\tabcolsep}{7pt}
    \caption{Ablation study of main components of our method.
    }
    \resizebox{\textwidth}{!}{
    \begin{tabular}{cccc|cc}
        \toprule
        
        Soft Cascade & Sub-Region-Aware Prompt & LoRA & Text Modulation & Dice (\%) $\uparrow$ & HD95 $\downarrow$ \\
        
        \midrule
        $\checkmark$ & $\times$ & $\times$ & $\times$ & 85.9 & 5.67 \\
        $\checkmark$ & $\checkmark$ & $\times$ & $\times$ & 86.4 & 5.26 \\
        $\checkmark$ & $\checkmark$ & $\checkmark$ & $\times$ & 86.6 & 5.04 \\
        $\checkmark$ & $\checkmark$ & $\checkmark$ & $\checkmark$ & \textbf{87.0} & \textbf{4.81} \\
        \bottomrule
    \end{tabular}}
    \vspace{-1mm}
    \label{tab::ablation_components}
\end{table}

\begin{table}[t]
    \centering
    \setlength{\tabcolsep}{1.5pt}
    \begin{minipage}{0.48\linewidth}
        \centering
        \caption{Ablation of cascading approach.
        }
        \vspace{-1mm}
        \begin{tabular}{cccccc}
            \toprule
            
            Cascade Approach & \multicolumn{4}{c}{Dice (\%) $\uparrow$} \\
        
            Approach & ET & WT & TC & Avg. \\
            
            \hline
        
            WT+TC+ET & 84.0 & 90.2 & 83.9 & 86.0 \\
        
            WT$\to$TC+ET & 85.0 & 90.5 & 84.8 & 86.7 \\
        
            WT$\to$TC$\to$ET & \textbf{85.3} & \textbf{90.7} & \textbf{ 85.1} & \textbf{87.0} \\
         
            \bottomrule
        \end{tabular}
        \label{tab::ablation_cascade}
    \end{minipage}
    \hfill
    \begin{minipage}{0.48\linewidth}
        \centering
        \caption{Ablation of learnable tokens.
        }
        \begin{tabular}{cccccc}
            \toprule
            
            \multicolumn{1}{c}{\multirow{2}{*}{\#Tokens}} & \multicolumn{4}{c}{Dice (\%) $\uparrow$} \\

            & ET & WT & TC & Avg. \\
            
            \hline
        
            1 & 84.1 & 90.3 & 84.2 & 86.2 \\
        
            4 & \textbf{85.3} & \textbf{90.7} & \textbf{85.1} & \textbf{87.0} \\
        
            10 & 84.8 & 90.5 & 84.7 & 86.7 \\
         
            \bottomrule
        \end{tabular}
        \label{tab::ablation_token}
    \end{minipage}
    \vspace{-4mm}
\end{table}

\subsection{Ablation Study}

\noindent \textbf{Effect of Components.}
Regarding Table~\ref{tab::ablation_components}We progressively add each proposed component to evaluate its individual contribution. Starting from the soft cascade alone ($85.9\%$ Dice), introducing sub-region-aware prompts results in a $+0.5\%$ improvement by enabling BioBERT to produce specialized text representations for each tumor sub-region. Adding LoRA further improves performance to $86.6\%$ Dice by allowing the text encoder to adapt its internal representations. Finally, incorporating the text-semantic channel modulators achieves the best result of $87.0\%$ Dice, as the channel-wise refinement allows the decoder to emphasize text-relevant feature channels at each cascade stage.

\noindent \textbf{Effect of Cascading Approach.}
We compare three cascade strategies in Table~\ref{tab::ablation_cascade}: independent parallel heads (WT+TC+ET), partial cascading where WT guides both TC and ET simultaneously (WT$\to$TC+ET), and our full sequential cascade (WT$\to$TC$\to$ET). Independent prediction performs worst ($86.0\%$), as each sub-region must learn its boundaries without spatial priors from coarser regions. Partial cascading improves the average to $86.7\%$; however, the full sequential cascade achieves the best results across all sub-regions ($87.0\%$).

\noindent \textbf{Effect of Number of Learnable Tokens.}
We vary the number of soft prompt tokens $K \in \{1, 4, 10\}$ prepended to the text input, as demonstrated in Table~\ref{tab::ablation_token}. A single token provides limited capacity to steer BioBERT toward sub-region-specific features. $K{=}4$ achieves the best performance, providing sufficient representational capacity for sub-region specialization. Increasing to $K{=}10$ slightly degrades performance, as the additional prompt tokens tend to wash out the word-level semantics during self-attention, consistent with the diminishing-return trend observed in the prompt tuning literature.

\begin{figure*}[t!]
    \centering
    \includegraphics[width=0.77\linewidth]{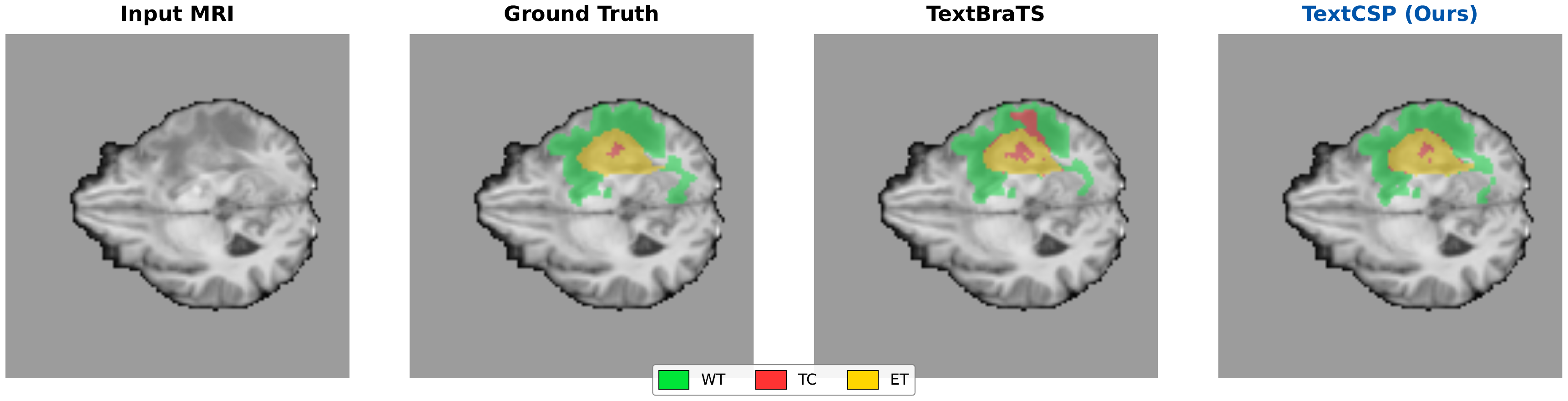}
    \vspace{-3mm}
    \caption{Qualitative comparison with the baseline TextBraTS.
    }
    \label{fig::qualitative_comparison}
\end{figure*}

\begin{figure*}[t!]
    \centering
    \includegraphics[width=0.8\linewidth]{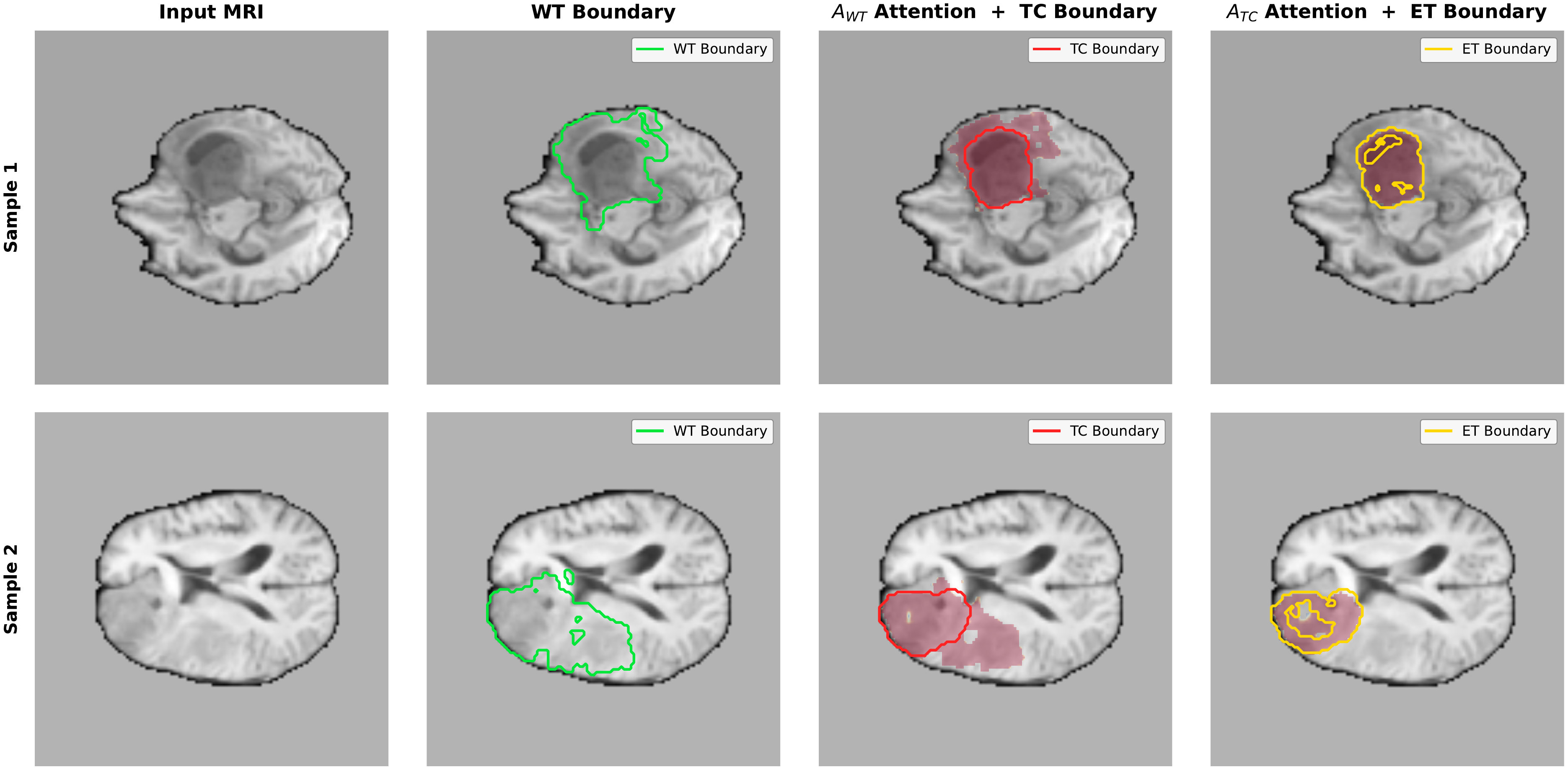}
    \vspace{-2mm}
    \caption{Cascade attention visualization. $A_{WT}$ concentrates within the WT region toward the TC boundary, and $A_{TC}$ further narrows from TC toward the ET boundary.
    }
    \label{fig::soft_attention}
    \vspace{-4mm}
\end{figure*}

\subsection{Qualitative Analysis}
Fig.~\ref{fig::qualitative_comparison} presents a qualitative comparison of segmentation results on a representative test case where TextCSP significantly outperforms our baseline. Moreover, Fig.\ref{fig::soft_attention} provides insight into why the cascade achieves this: the WT attention map $A_{WT}$ activates broadly across the whole tumor and concentrates toward the TC boundary, while the TC attention map $A_{TC}$ further narrows its focus from the tumor core to tightly encompass the ET boundary. This learned coarse-to-fine spatial refinement, following the anatomical hierarchy WT$\to$TC$\to$ET, enables each cascade stage to provide a meaningful spatial prior for the subsequent finer sub-region without requiring any intermediate supervision.

\vspace{-1mm}
\section{Conclusion}
\vspace{-2mm}
We present TextCSP, a text-guided brain tumor segmentation framework that introduces soft cascade decoder to exploit the anatomical hierarchy of tumor sub-regions, and sub-region-aware prompt tuning. By prepending learnable prompts to BioBERT with lightweight LoRA adaptation, our method generates specialized text representations for each sub-region. The soft cascade progressively refines spatial attention from WT to TC and then to ET, providing each stage with an anatomical prior from its parent region. Text-semantic channel modulators further bridge the text and visual features through channel-wise refinement. Extensive experiments on the TextBraTS dataset demonstrate that TextCSP achieves SOTA performance, and we show that all components are effective.

\bibliographystyle{splncs04}
\bibliography{ref}

@inproceedings{xin2025text3dsam,
  title={Text3DSAM: Text-Guided 3D Medical Image Segmentation Using SAM-Inspired Architecture},
  author={Xin, Yu and Ates, Gorkem Can and Shao, Wei},
  booktitle={CVPR 2025: Foundation Models for 3D Biomedical Image Segmentation},
  year={2025}
}

@misc{hu2019squeezeandexcitationnetworks,
      title={Squeeze-and-Excitation Networks}, 
      author={Jie Hu and Li Shen and Samuel Albanie and Gang Sun and Enhua Wu},
      year={2019},
      eprint={1709.01507},
      archivePrefix={arXiv},
      primaryClass={cs.CV},
}

@article{xie2026tvpnet,
  title={TVPNet: Text-vision prompt guided segmentation for small 3D medical object},
  author={Xie, Yuhua and Wang, Rui and Zhuang, Yan and Chen, Ke and Han, Lin and Liao, Guoliang and Hou, Yao and Hou, Linxuan and Lin, Jiangli},
  journal={Biomedical Signal Processing and Control},
  volume={111},
  pages={108239},
  year={2026},
  publisher={Elsevier}
}

@article{rokuss2025voxtell,
  title={Voxtell: Free-text promptable universal 3d medical image segmentation},
  author={Rokuss, Maximilian and Langenberg, Moritz and Kirchhoff, Yannick and Isensee, Fabian and Hamm, Benjamin and Ulrich, Constantin and Regnery, Sebastian and Bauer, Lukas and Katsigiannopulos, Efthimios and Norajitra, Tobias and others},
  journal={arXiv preprint arXiv:2511.11450},
  year={2025}
}

@inproceedings{shi2025textbrats,
  title={TextBraTS: Text-Guided Volumetric Brain Tumor Segmentation with Innovative Dataset Development and Fusion Module Exploration},
  author={Shi, Xiaoyu and Jain, Rahul Kumar and Li, Yinhao and Hou, Ruibo and Cheng, Jingliang and Bai, Jie and Zhao, Guohua and Lin, Lanfen and Xu, Rui and Chen, Yen-wei},
  booktitle={International Conference on Medical Image Computing and Computer-Assisted Intervention},
  pages={638--648},
  year={2025},
  organization={Springer}
}

@inproceedings{luo2025vividmed,
  title={Vividmed: Vision language model with versatile visual grounding for medicine},
  author={Luo, Lingxiao and Tang, Bingda and Chen, Xuanzhong and Han, Rong and Chen, Ting},
  booktitle={Proceedings of the 2025 Conference of the Nations of the Americas Chapter of the Association for Computational Linguistics: Human Language Technologies (Volume 1: Long Papers)},
  pages={1800--1821},
  year={2025}
}

@article{liu2025medsam3,
  title={MedSAM3: Delving into Segment Anything with Medical Concepts},
  author={Liu, Anglin and Xue, Rundong and Cao, Xu R and Shen, Yifan and Lu, Yi and Li, Xiang and Chen, Qianqian and Chen, Jintai},
  journal={arXiv preprint arXiv:2511.19046},
  year={2025}
}

@inproceedings{hatamizadeh_2021_swin,
  title={Swin unetr: Swin transformers for semantic segmentation of brain tumors in mri images},
  author={Hatamizadeh, Ali and Nath, Vishwesh and Tang, Yucheng and Yang, Dong and Roth, Holger R and Xu, Daguang},
  booktitle={International MICCAI brainlesion workshop},
  pages={272--284},
  year={2021},
  organization={Springer}
}

@inproceedings{hsu_2021_brain,
  title={Brain tumor segmentation (BraTS) challenge short paper: Improving three-dimensional brain tumor segmentation using SegResNet and hybrid boundary-dice loss},
  author={Hsu, Cheyu and Chang, Chunhao and Chen, Tom Weiwu and Tsai, Hsinhan and Ma, Shihchieh and Wang, Weichung},
  booktitle={International MICCAI Brainlesion Workshop},
  pages={334--344},
  year={2021},
  organization={Springer}
}

@article{isensee_2021_nnu,
  title={nnU-Net: a self-configuring method for deep learning-based biomedical image segmentation},
  author={Isensee, Fabian and Jaeger, Paul F and Kohl, Simon AA and Petersen, Jens and Maier-Hein, Klaus H},
  journal={Nature Methods},
  volume={18},
  number={2},
  pages={203--211},
  year={2021},
  publisher={Nature Publishing Group US New York}
}

@InProceedings{olaf_2015_unet,
author="Ronneberger, Olaf
and Fischer, Philipp
and Brox, Thomas",
editor="Navab, Nassir
and Hornegger, Joachim
and Wells, William M.
and Frangi, Alejandro F.",
title="U-Net: Convolutional Networks for Biomedical Image Segmentation",
booktitle="MICCAI",
year="2015",
publisher="Springer International Publishing",
pages="234--241"
}

@inproceedings{xing_2022_nestedformer,
  title={NestedFormer: Nested modality-aware transformer for brain tumor segmentation},
  author={Xing, Zhaohu and Yu, Lequan and Wan, Liang and Han, Tong and Zhu, Lei},
  booktitle={MICCAI},
  pages={140--150},
  year={2022},
  organization={Springer}
}

@article{dice_1945_measures,
  title={Measures of the amount of ecologic association between species},
  author={Dice, Lee R},
  journal={Ecology},
  volume={26},
  number={3},
  pages={297--302},
  year={1945},
  publisher={JSTOR}
}

@article{huttenlocher_2002_comparing,
  title={Comparing images using the Hausdorff distance},
  author={Huttenlocher, Daniel P and Klanderman, Gregory A. and Rucklidge, William J},
  journal={IEEE Transactions on pattern analysis and machine intelligence},
  volume={15},
  number={9},
  pages={850--863},
  year={2002},
  publisher={IEEE}
}

@article{paszke_2017_pytorch,
  title={Automatic differentiation in pytorch},
  author={Paszke, Adam and Gross, Sam and Chintala, Soumith and Chanan, Gregory and Yang, Edward and DeVito, Zachary and Lin, Zeming and Desmaison, Alban and Antiga, Luca and Lerer, Adam},
  year={2017}
}

@article{cardoso_2022_monai,
  title={Monai: An open-source framework for deep learning in healthcare},
  author={Cardoso, M Jorge and Li, Wenqi and Brown, Richard and Ma, Nic and Kerfoot, Eric and Wang, Yiheng and Murrey, Benjamin and Myronenko, Andriy and Zhao, Can and Yang, Dong and others},
  journal={arXiv preprint arXiv:2211.02701},
  year={2022}
}

@inproceedings{foret_2021_sam,
  title={Sharpness-aware Minimization for Efficiently Improving Generalization},
  author={Pierre Foret and Ariel Kleiner and Hossein Mobahi and Behnam Neyshabur},
  booktitle={International Conference on Learning Representations},
  year={2021}
}

@inproceedings{nesterov_1983_sgd,
  title={A method for solving the convex programming problem with convergence rate O (1/k2)},
  author={Nesterov, Yurii},
  booktitle={Dokl akad nauk Sssr},
  volume={269},
  pages={543},
  year={1983}
}

@inproceedings{hu_2021_lora,
title={Lo{RA}: Low-Rank Adaptation of Large Language Models},
author={Edward J Hu and Yelong Shen and Phillip Wallis and Zeyuan Allen-Zhu and Yuanzhi Li and Shean Wang and Lu Wang and Weizhu Chen},
booktitle={ICLR},
year={2022}
}

@article{lee_2020_biobert,
  title={BioBERT: a pre-trained biomedical language representation model for biomedical text mining},
  author={Lee, Jinhyuk and Yoon, Wonjin and Kim, Sungdong and Kim, Donghyeon and Kim, Sunkyu and So, Chan Ho and Kang, Jaewoo},
  journal={Bioinformatics},
  volume={36},
  number={4},
  pages={1234--1240},
  year={2020},
  publisher={Oxford University Press}
}

\end{document}